# VidCEP: Complex Event Processing Framework to Detect Spatiotemporal Patterns in Video Streams

Piyush Yadav
Lero- Irish Software Research Centre
*National University of Ireland Galway*
Galway, Ireland
piyush.yadav@lero.ie

Edward Curry
Lero- Irish Software Research Centre
*National University of Ireland Galway*
Galway, Ireland
edward.curry@lero.ie

*Abstract*— Video data is highly expressive and has traditionally been very difficult for a machine to interpret. Querying event patterns from video streams is challenging due to its unstructured representation. Middleware systems such as Complex Event Processing (CEP) mine patterns from data streams and send notifications to users in a timely fashion. Current CEP systems have inherent limitations to query video streams due to their unstructured data model and lack of expressive query language. In this work, we focus on a CEP framework where users can define high-level expressive queries over videos to detect a range of spatiotemporal event patterns. In this context, we propose- i) VidCEP, an in-memory, on the fly, near real-time complex event matching framework for video streams. The system uses a graph-based event representation for video streams which enables the detection of high-level semantic concepts from video using cascades of Deep Neural Network models, ii) a Video Event Query language (VEQL) to express high-level user queries for video streams in CEP, iii) a complex event matcher to detect spatiotemporal video event patterns by matching expressive user queries over video data. The proposed approach detects spatiotemporal video event patterns with an F-score ranging from 0.66 to 0.89. VidCEP maintains near real-time performance with an average throughput of 70 frames per second for 5 parallel videos with sub-second matching latency.

I. INTRODUCTION

Complex Event Processing (CEP) systems have been increasingly adopted for real-time analysis in different domains, such as traffic and financial applications [1], [2]. CEP systems combine individual atomic events from streams to form meaningful high-level semantics and notify interested users in a timely manner [2], [3]. The key characteristics of CEP systems are i) easily *express* event patterns of interest and ii) *detect* event patterns in near *real-time* by performing *matching* over streams.

With the recent advancement in digital and sensor technology, there is a significant shift in the nature of data streams. Visual sensors like smartphones and cameras are ubiquitous and are generating an unprecedented amount of video data. For example, cities like London and New York have deployed thousands of CCTV cameras, streaming hours of videos daily [4]. Analytics is performed over these video streams to detect events of interest in applications like business intelligence, surveillance, and traffic monitoring [5].

Video streams are highly rich in semantic information, but it is challenging to detect event patterns from them because of their low-level features (like pixels). Current CEP systems have limitations to detect event patterns over videos [6]. For example, a user may be interested in analyzing the interaction between 'Car' and 'Bike' events in five-minute intervals in a video and issue a 'Car AND Bike' query to detect video frames related to such spatiotemporal event patterns. The processing of video queries in CEP leads to challenges including:

- How to define high-level human-understandable *expressive video event pattern queries* like 'Car AND Bike' in CEP?
- How to *write event rules* for 'Car AND Bike' which occurs over space and time?
- How to *match* low-level video data with high-level declarative queries in a CEP system?

There is a rich body of work for querying and analyzing video content in the database community [7]–[11] but less attention has been paid to content-based event pattern detection of video streams in CEP systems [12]. There is a need to overcome the challenges like matching, querying and event rules to enhance the CEP systems with video processing capabilities.

We propose a video stream enabled CEP system with the following contributions:

1. *Architecture:* We introduce a complex event processing framework-VidCEP which can handle multiple parallel video streams and perform continuous pattern matching in near real-time.

2. *Event Query Language:* We present a high-level Video Event Query Language (VEQL) to query events from the video stream. VEQL enables the user to write expressive, actionable and explicit CEP queries for video data with different spatial and temporal constructs.

3. *Event Matching*: We propose a method for the content-based event matching using VEQL to perform state-based spatiotemporal event matching. The matching execution model is continuous, where event patterns are detected in near real-time.

The rest of the paper is organized as follows: Section II explains background and motivation, Section III introduces the proposed approach, Section IV formalizes VEQL query constructs, Section V describes the VidCEP architecture while Section VI focuses on event matching. The experimental results, related work and conclusion are explained in Section VII, Section VIII and Section IX, respectively.

This work was supported with the financial support of the Science Foundation Ireland grant 13/RC/2094 and co-funded under the European Regional Development Fund through the Southern & Eastern Regional Operational Programme to Lero - the Irish Software Research Centre (www.lero.ie)

## II. BACKGROUND AND MOTIVATION

### A. Image Understanding

The image understanding domain focuses on reasoning over image content and describes the image using high-level human-understandable concepts. In computer vision, these high-level visual concepts are termed as '*Objects*'. Objects are the basic building block of images which are a collection of low-level features like 'pixels', 'intensity', 'color', 'edges' and have been given high-level semantic labels like 'Car', and 'Bike'. Algorithms from vision literature (like SIFT [13]) can detect objects from the images. Recently, Deep Neural Networks (DNN) [14] have become a state-of-the-art method to identify objects with good accuracy and performance. DNN-based object detection models like YOLO [15], and M-RCNN [16] give bounding boxes around the objects in the images which are highly accurate.

TABLE 1. QUERY DIMENSIONS FOR CEP VIDEO PROCESSING

| Query Dimensions | Description | User Query Example |
|---|---|---|
| **D1:** Object Detection | The user wants to detect an object from the video stream. | Notify if any '*Car*' is present in the video feed. |
| **D2:** Object Detection with Specific Attributes | The user wants to detect an object with specific attributes (like color, type) from the video stream. | Notify if any '*Red Car*' is present in the video feed. |
| **D3:** Spatial Relationship among Objects | The user wants to detect an event pattern where two objects are related spatially. This can be object location or direction with respect to other objects. | Notify if any '*Car*' is present on the '*Left*' side of another '*Car*'. |
| **D4:** Temporal Relationship among Objects | The user wants to query objects which are related temporally. This can be in term of occurrence of objects which are arranged in some temporal fashion. | Notify if '*Car*' appears before '*Truck*'. |

### B. Motivational Scenario

Objects in videos interact in both space and time and require spatiotemporal reasoning to query them. We have divided the video query requirements for CEP into four dimensions- D1- Object detection, D2- Object detection with specific Attributes, D3- Spatial Relationships among objects, and D4- Temporal Relationships among objects. Table 1 maps the above video query requirements and enlists events which users might be interested in querying from video streams. Suppose in a smart city the traffic control authority is searching for a 'Red Car' on a specific road section. They have subscribed to camera video feeds of the road to get automated real-time notifications for detection of 'Red Car' in a 10-second interval. Here a 'Red Car' event needs to be continuously monitored by the CEP engine to detect an object 'Car' with a specific attribute (color- Red). Fig. 1 shows different event patterns using these query dimensions. If we analyse different video frame sequences, then the 'Red Car' event occurs in the video at time t1 (D2). At time t3 there is a spatial event where 'Car 1' spatially occurs 'left' of 'Car 2' (D3). The temporal event where 'Car' occurs before 'Truck' is at time [t1, t2] where t1< t2 (D4). Similarly, the traffic control authority may want to monitor high traffic flow on the road at a given time of the day. Here a 'High Traffic Flow' event is composed of simple events like– a) Detection of 'Car' events (D1), and b) Counting the number of cars in each frame at different time instances (D4). In Fig. 1, there was less traffic until time t3 but 'high traffic flow' at time t4. If a user wants to query for these situations in present CEP systems, they will encounter a number of challenges which are discussed in Section II-D.

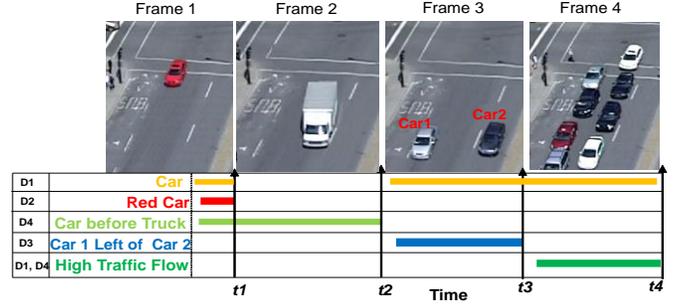

Fig. 1. Spatiotemporal event patterns in a video

### C. Spatiotemporal Support for Video Query in Event Query Languages

CEP systems have their formalism and syntax to query events using Event Query Languages (EQL) [17]–[27]. Most of the present EQL queries are expressed using high-level SQL-like declarative language that define complex event patterns using event rules. EQL perform operations like filter, aggregate, windows to detect simple atomic events and correlate them to more high-level meaningful information in data streams.

TABLE 2: COMPARISON OF EQL AND VIDEO QUERY LANGUAGES

| | Support Object Detection (D1) | Support Objects and Attributes Detection (D2) | Support Spatial Relations (D3) | Support Temporal Relations (D4) | Support Video Stream Processing |
|---|---|---|---|---|---|
| **Event Query Languages** | | | | | |
| Snoop [16] | No | No | No | Yes | No |
| CEDR [17] | No | No | No | Yes | No |
| SASE [19] | No | No | No | Yes | No |
| Xchange$^{EQ}$ [20] | No | No | No | Yes | No |
| Esper EPL [21] | No | No | Partial | Yes | No |
| StreamSQL [22] | No | No | No | Yes | No |
| TESLA [23] | No | No | No | Yes | No |
| Cayuga [24] | No | No | No | Yes | No |
| EP-SPARQL [25] | No | No | No | Yes | No |
| SPARQL-ST [26] | No | No | Yes | Yes | No |
| SPARQL-MM [18] | No | No | Yes | Yes | No |
| **Video Query Languages in Databases** | | | | | |
| VSQL [27] | Yes | Yes | Yes | No | No |
| VISUAL [28] | Yes | Yes | Partial | No | No |
| MOQL [10] | Yes | No | Partial | Partial | No |
| FRAMEQL [18] | Yes | Yes | No | No | No |
| Le et.al. [48] | Yes | No | Partial | No | No |
| CVQL [30] | Yes | No | Partial | No | No |
| VERL [31] | Yes | Partial | Partial | Yes | No |
| VIQS [32] | Yes | No | Partial | Partial | No |
| BilVideo [33] | Yes | Yes | Yes | Yes | No |
| SVQL [6] | Yes | Yes | Yes | Yes | No |

Since the '90s, there have been many proposals for video query languages in databases [7], [9], [11], [28]–[34]. These queries extract pre-annotated video content indexed in databases using fixed schemas. Table 2 shows the descriptive comparison of current EQL with the video query languages in databases. It is evident from the table that no EQL fully supports the processing of video streams as per the identified query dimensions, and they mostly focus on temporal reasoning. SPARQL-MM [19] is an EQL with some video querying capabilities, but it deals with linked video data where objects coordinates in frames are pre-annotated. FRAMEQL [9] has been proposed as a SQL-like query language that uses DNNs to answer relational queries over video content. FRAMEQL focuses on creating a system with database functionality as opposed to supporting CEP with spatial and temporal event patterns.

## D. Challenges

Fig. 2 shows a comparison between the present CEP scenario and the functionalities required to process video streams. As shown in Fig. 2, a CEP system can be divided into three broad functional components [35]: 1) *Event Representation*- defines a data model for the incoming stream on the basis of a structured schema, 2) *Event Query language*- queries events of interest using EQL, and 3) *Event matching*- detects event patterns based on registered queries. The CEP system receives data streams from producers (source) and sends an event notification to consumers (users) who have queried for interesting events. Below we list the challenges that the CEP system requires to process video data.

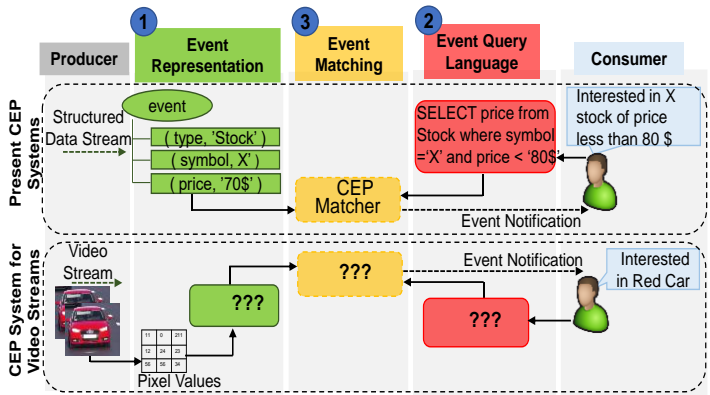

Fig. 2. Challenges in CEP to process video streams

*Unstructured Event Representation:* Present CEP engines like Esper [22] and Cayuga [25] assume that incoming data streams have a fixed data model with a structured payload such as key-value pairs, XML data and RDF triples [36]. Reasoning over a structured data model with well-defined semantics and representation is a well-known problem. As shown in Fig. 2, the video is represented as low-level features (pixels values) while humans interpret the video content as high-level semantic labels ('Red Car'). Current CEP systems are not capable of handling video streams since the low-level features of images do not match to the high-level features of a query.

*Query expression for video event patterns in CEP:* EQL performs relational operations to extract information from structured data streams. In Fig. 2, a user interested in 'X' Stock with a value less than $80 can easily express their interest using SQL-like declarative syntax. Writing expressive and declarative queries like 'Red Car' for low-level video content (pixel values) is challenging. There is a need to develop an expressive, spatial and temporal reasoning-based query mechanism to support pattern detection for videos in CEP.

*Complex Video Event Matching:* Video streams have evolving nature where objects are in motion and generate varying complex patterns. These patterns are spread across spatial and temporal dimensions. For example, a 'Car' can be present over time, but it occupies a specific space in the video, which can change. This adds an extra layer of complexity to match the event patterns both at the temporal and spatial levels. Existing CEP systems mostly focus on temporal pattern matching in data streams (Fig. 2). There is a need to build video event matching capability in CEP, which can perform matching both at the spatial and temporal levels.

## III. APPROACH

### A. Video Event Definition

In CEP, an *event* is considered as an occurrence that has happened in a domain [37]. The *video event* is a high-level semantic concept captured due to the change of state in video content while observed over time [32]. Using CEP analogy, two categories of video events are defined [38].

*Simple Video Event:* A simple (atomic) event is the instantaneous (i.e. either exists entirely or not at all) happening of interest [20] and is directly detectable. *Objects* are the primary visual concepts which user perceive from a video. A Simple Video Event can be considered as an occurrence of an object in the video. For example, if a user is interested in specific objects (such as 'Cat', 'Bus') in a video, then it is considered as a Simple Video Event.

*Complex Video Event:* Complex events are *composed* events which are derived from aggregating simple events [3]. The complexity of the events depends on the application logic where simple events (from the same or different streams) are nested with temporal and logical operators to form a complex event. A Complex Video Event is composed of simple video events using spatiotemporal operators. For example, the presence of 'Car AND Bike' is a complex video event derived from the simple video events, i.e. 'Car' and 'Bike'.

### B. Video Event Representation

Videos streams are ordered sequence of image frames where each frame represents a data item. Image frames are represented as low-level features with no fixed data model. These frames need to be converted into a structured representation which can act as input to the CEP engine. From the representation perspective, a video can be divided into two aspects: 1) Objects, and 2) Relationships among Objects [38].

*Objects:* A machine interprets image frame as low-level visual features (e.g. pixels and edges) while users perceive them as high-level semantic concepts, i.e. Objects (e.g. 'Person'). These objects can have different properties which are represented by their attributes (e.g. color, shape, type). The representation should handle video objects at various levels ranging from objects and attributes to low-level features.

*Relationships among Objects:* In a video, relationships among objects can exist within a frame (intraframe) and across frames (interframe). Within an image frame, objects interact with each other spatially and occupy specific positions. Across frames (interframe), objects interact with each other temporally. Thus, a suitable representation should be able to handle the object's spatiotemporal information at the frame level (intraframe) and stream level (interframe).

The above aspects have been mapped to an event-centric structured representation. In our previous work [38], we proposed a semantic event representation of video streams in the form of graphs using Video Event Knowledge Graph (VEKG). In VEKG, the low-level video data is extracted and represented to a structured data model.

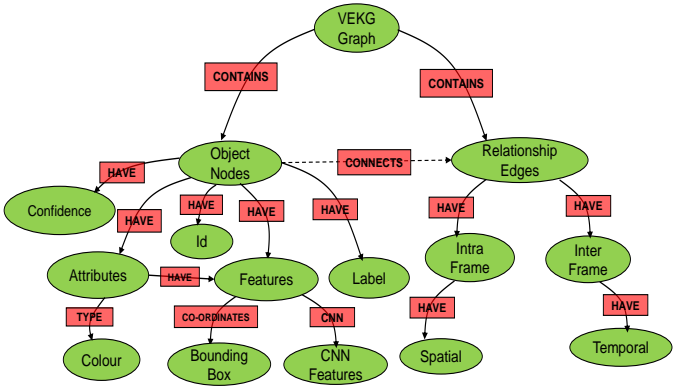

Fig. 3. Video Event Knowledge Graph (VEKG) schema [38]

Fig. 3 shows the Video Event Knowledge Graph (VEKG) representation, where nodes correspond to objects and edges represent spatial and temporal relationships among objects. Thus, VEKG can be defined as:

***Definition1 (VEKG Graph):*** For any image frame, the resulting Video Event Knowledge Graph is a labelled graph with six tuples represented as VEKG = {$V, E, A_V, R_E, \lambda_V, \lambda_E$} where

$V$ = set of object nodes $O_i$

$E$ = set of edges such $E \subseteq V \times V$

$A_V$ = set of properties mapped to each object nodes such that $O_i$ = (id, attributes, label, confidence, features)

$R_E$ = set of spatiotemporal relation classes $\lambda_V, \lambda_E$ are class labelling functions $\lambda_V: V \rightarrow O$ and $\lambda_E: E \rightarrow R_E$

***Definition2 (VEKG Stream):*** A Video Event Knowledge Graph Stream is a sequence ordered representation of VEKG such that $VEKG(S) = \{(VEKG_1, t_1), (VEKG_2, t_2) \ldots (VEKG_n, t_n)\}$ where $t \epsilon\ timestamp$ such that $t_i < t_{i+1}$.

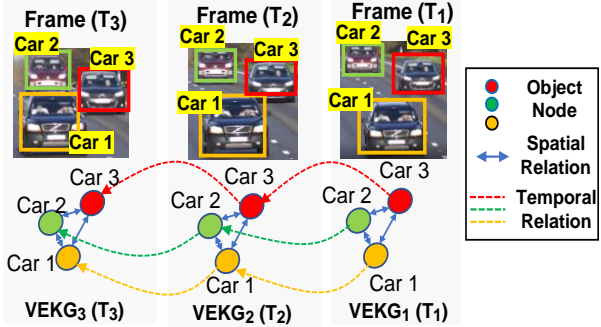

Fig. 4. VEKG graph for each video frame [38]

Fig. 4 shows a VEKG stream of three VEKG graphs for image frames at different time instances. The object nodes (Car1, Car2, Car3) in VEKG graphs are connected using spatial and temporal edges. VEKG is a complete digraph, which means that each object is spatially related to another object which is present in the image frame. Initially, the edge weight is set to zero, and the weights between nodes are updated during matching as per query. The temporal relation edge between object nodes is created by identifying the same object nodes in different frames using object tracking. The tracking is performed based on cosine distance between objects features and Intersection Over Union (IOU) [15] of bounding boxes which are extracted using DNN models.

## IV. VEQL: A COMPLEX EVENT QUERY LANGUAGE FOR VIDEO STREAMS

### A. Basic Query Syntax

For video event detection in CEP, there is a requirement of query expression which enables the user to query video events in high-level human-understandable concepts without worrying about its low-level features. As discussed earlier, most of the existing CEP languages use SQL-like declarative languages and support predefined basic operators. We introduce the Video Event Query Language (VEQL), which follows the SQL-like declarative expression. The aim is that by using a standardized vocabulary of existing event query languages, it will be easier for CEP users to express precise video queries and integrate video events information seamlessly. The query syntax of VEQL is:

***SELECT* < *pattern* > *FROM* < *producer* > *WHERE* < *condition* > *WITHIN* < *window* > *WITH_CONFIDENCE* < *confidence score* >**

In the above VEQL syntax- 1) *pattern* refers to the desired event of interest, 2) *producer* refers to video streaming source, 3) *condition* is specific constraints which event predicates must satisfy, 4) *windows* refers to interval of a stream, i.e. state over which matching is performed, and 5) *confidence score* is a threshold matching score [0-1] which the consumer will accept as a result. We now explain the different query dimensions which we have enlisted in our motivation through various VEQL query examples. The query patterns are listed as per their increasing complexity.

**D1-Object Detection:** Query1 (Q1) retrieves the object of interest from the video stream. Q1 is the simplest video event pattern query where the *pattern* clause is an 'Object'. The WHERE clause filters the object based on query predicate 'label' which is a 'Car'. The TIMEFRAME_WINDOW clause specifies the detection of an object within a time window of 10 seconds. This means that the window will accept the VEKG stream for 10 seconds. The WITH_CONFIDENCE clause determines the minimum matching confidence of the object, which is 0.5.

```
Q1: SELECT Object FROM Camera
    WHERE Object.label= 'Car'
    WITHIN TIMEFRAME_WINDOW(10) WITH_CONFIDENCE > 0.5
```

**D2-Object Detection with Specific Attributes:** Query 2 (Q2) is like Q1 with the addition of specific object attributes. Here the WHERE clause consists of two predicates- an object label which is a 'Car' and an object attribute which is of color 'Black'. It is worth mentioning that the WHERE clause, in general, can be a combination of logical operators (AND, OR) and comparison operators (<, >, =, ≤, ≥).

```
Q2: SELECT Object FROM Camera
    WHERE Object.label='Car'
    AND Object.attrcolor = 'Black'
    WITHIN TIMEFRAME_WINDOW(10) WITH_CONFIDENCE > 0.5
```

**D3-Spatial Relationship among Objects:** The third VEQL query (Q3) identifies the spatial relationship between two objects. In the *pattern* clause of Q3, 'Left' is one of the spatial relations which need to be established between two objects, i.e. 'Car' with color attributes- 'Black' and 'Not Black'. Q3 identifies an event pattern where a 'Black Car' is present on the *left* of the 'Not Black' car in a time window of 10 seconds.

```
Q3: SELECT Left(Object1, Object2)FROM Camera
    WHERE Object1.label= 'Car' AND
    Object1.attrcolor = 'Black' AND
    Object2.label = 'Car' AND
    Object2.attrcolor = 'Not Black'
    WITHIN TIMEFRAME_WINDOW(10) WITH_CONFIDENCE > 0.5
```

**D4-Temporal Relationship among Objects:** Query 4 (Q4) establish a temporal relationship between different objects. EQL's like Snoop [17], CEDR [18], and SASE [20] have well-established composition operators which support temporal relationships. VEQL adopts temporal constructs from these languages for the processing of video streams. In Q4 the *pattern* clause sequence (SEQ) identifies the occurrence of objects in a time window. Query 4 enlists that two objects 'Car' and 'Person' are occurring in a sequence within 10 seconds.

```
Q4: SELECT SEQ(Object1, Object2) FROM Camera
    WHERE Object1.label= 'Car'
    AND Object2.label = 'Person'
    WITHIN TIMEFRAME_WINDOW(10) WITH_CONFIDENCE > 0.5
```

**D1, D4-Count Objects across Frames:** In query 5 (Q5) the *pattern* clause detects a 'High Traffic Flow' event. The WHERE clause has two predicates to define high traffic flow event. The COUNT of an object, i.e. 'Car' should be greater than a threshold (here 5) and the count of objects should be consistent across each frame which is denoted by FOR EACH FRAME clause. Thus, Q5 is a combination of detecting objects and counting them in each frame to detect the high or low traffic flow pattern. In Q5, HIGH_TRAFFIC_FLOW is a context-specific example and gives an idea of how VEQL can be used in different domains.

```
Q5: SELECT HIGH_TRAFFIC_FLOW(Object)FROM Camera
    WHERE Object.label= 'Car'
    AND COUNT(Object)> 5 FOR EACH FRAME
    WITHIN TIMEFRAME_WINDOW(10) WITH_CONFIDENCE > 0.5
```

The above VEQL queries are converted to a query graph which is used for graph-based matching over VEKG streams.

### B. Formalizing Spatial and Temporal Constructs for VEQL

*1) Spatial Built-in Conditions:* The interaction among objects occurs in a spatial dimension in which relationship can be modelled using spatial relations. We have modelled spatial relations (S) into three main classes:

- *Geometric Representation for Spatial Object (O):* Current deep learning techniques identify objects from video data by either creating bounding boxes [15] or by creating segmented region across these objects' boundaries [16]. For spatial interaction, bounding box based geometry is considered to represent objects (Fig. 5).
- *Direction Based Spatial Relation ($S_D$):* Direction captures the projection and orientation of an object in space. We have used the simpler version of the Fixed Orientation Reference System (FORS) [39] and divided the space into four regions: *{front, back, left, right}* as shown in Fig. 5.
- *Topology-Based Spatial Relation ($S_T$):* We used Dimensionally Extended Nine-Intersection Model (DE-9im), a 2-dimensional topological model which describes a pairwise relationship between spatial geometries (O). This model has nine relationships: {*Disjoint, Touch, Contains, Intersect, Within, Covered by, Crosses, Overlap, Inside*} of which four relations are shown in Fig. 5.

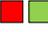

Fig. 5. Spatial relations

*2) Spatial Functions:* To calculate the spatial edge weight in VEKG, we have devised two types of spatial functions:

- *Boolean Spatial Function (bsf):* It returns the boolean weight between spatial objects (O). For example, for a direction relation ($S_D$) '*Left*', the boolean spatial function for two objects $O1$, and $O2$ will be *bsf (Left (O1, O2)) = 0 or 1*.
- *Metric Spatial Function (msf): msf* calculates the numerical (real number) weight between the spatial objects. We have defined two metric spatial functions- 1) *DISTANCE* that calculates the distance between two spatial objects and 2) *COUNT* that calculates the number of objects in an image frame.

*3) Temporal Built-in Conditions:* We have used the Allen Interval Algebra [40] to define four temporal patterns.

- *SEQ:* A sequence temporal relation can be defined when events occur in increasing chronological order. It can be expressed as $SEQ(E_1, E_2, ... E_n)$ when $E_1.t_1 < E_2.t_2 < ... < E_n.t_n$ where $E_i \in simple\ video\ event$ and $t \in time$. Here, time is taken as discrete and arranged in linear order $\{t_1, t_2, t_3 ...\}\ where\ t_i < t_{i+1}$.
- *EQ:* Equality is a concurrent relationship when two events occur at the same time instance. The temporal pattern $EQ(E_1, E_2, ... E_n)$ holds when $E_1.t_1 = E_2.t_2 = ...... = E_n.t_n$.
- *CONJ:* CONJ refers to the AND condition. $CONJ(E_1, E_2, ...... E_n)$ means that 'all' events should occur but there is no time order of their occurrence.
- *DISJ*: DISJ refers to the OR condition. $DISJ(E_1, E_2, ...... E_n)$ condition holds if 'any' of the given events occur irrespective of their time occurrence.

Only basic spatiotemporal operators have been defined to show the initial efficacy of the VidCEP. More complex operators will be addressed in future work.

## V. VidCEP ARCHITECTURE

Fig. 6 describes the architecture of the VidCEP engine. The engine architecture is divided into three main components: 1) Event Representation, 2) Event Matching, and 3) Event Query.

### A. Event Representation Component

The event representation component consists of *video stre-*

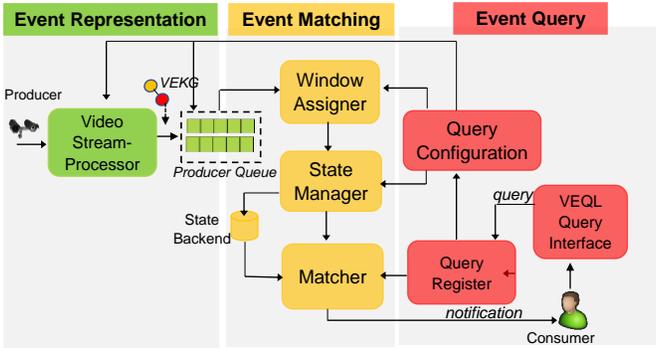

Fig. 6. VidCEP architecture

*-am processor* and a *producer queue*. A video stream processor is a computer vision pipeline that receives the video streams from different producers and converts them into a stream of VEKG graphs (cf. Fig. 7(a)). It consists of:

- *Video Frame Decoder*: This component receives the video frames and converts them to a low-level feature map.
- *DNN Models*: It constitutes different DNN models cascade (object detectors, attribute classifiers) pre-trained on specific datasets. The low-level feature map from the video decoder is passed to the object detector for detecting objects. The region of interest (ROI) of detected object features are then passed to attribute classifier for attribute detection.
- *Graph Constructor:* This module constructs a timestamped graph snapshot. It receives information from the DNN models and represents them as a graph based on the VEKG schema. VEKG graphs are pushed to the *producer queue,* which buffers the incoming streams and sends them to window assigner component for further processing.

### B. Event Matching Component

In CEP, *windows* capture the *state* and apply event rules to detect patterns over that state [41]. In our case windows capture the number of image frames of video streams which are represented as VEKG graphs. We have defined windows as:

$$[p]TIMEFRAME\_WINDOW\ (VEKG(S), t): \rightarrow S' \quad (1)$$

As per eq. 1, $TIMEFRAME\_WINDOW$ is applied over an incoming $VEKG(S)$ stream and gives a fixed subsequence $S' = ((VEKG_1, t_1), (VEKG_2, t_2) \ldots \ldots (VEKG_3, t_n))$ based on parameter $p$. The parameter $p$ is defined based on the sliding and tumbling nature of the window [42]. In eq. 1, time is discrete and arranged in a chronological order $\{t_1, t_2, t_3 \ldots \ldots\}$ where $t_i < t_{i+1}$. The event matching consists of the following components:

- *Window Assigner:* It assigns windows to different video streams as per the query. The window captures video frames as VEKG graphs into a fixed bucket size, i.e. 'state' and then applies a trigger function [42]. The trigger function is based on parameter $p$, which sends the captured state to the state manager when $p$ is satisfied. For example, in the VEQL queries, we have used TIMEFRAME_WINDOW(10), which means that the window captures the video stream for 10 seconds after which a trigger function will be activated.

- *State Manager*: The state manager receives the window state from different producers and sends them to the state backend and matcher. The state backend is a persistent storage which stores the event state of producers for historical analysis. The state manager is responsible for sending the state to the specific matcher instance for which the query was registered. If the matcher instance is busy, the state manager buffers the state and keeps it in a queue and waits until the matcher becomes free. At present, we are focusing on the state which is sent directly to the matcher.

- *Matcher:* The matcher (cf. Fig. 7(b)) receives the window state from state manager and applies spatial and temporal operations over it as discussed in Section VI.

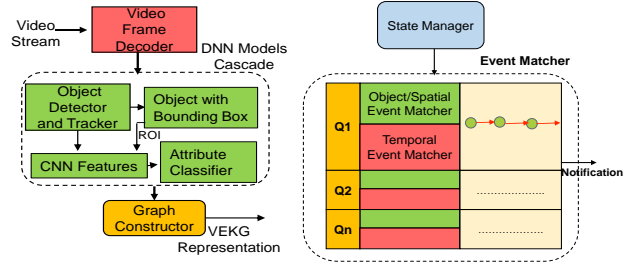

Fig. 7. (a) Video stream-processor and (b) Event matcher

### C. Event Query

The event query constitutes of three basic components:

- *VEQL Query Interface:* This is a user interface where users can write VEQL queries. We have already discussed a set of queries to detect different video patterns.
- *Query Register:* Acts as a registry where all continuous queries from different consumers are indexed.
- *Query Configuration:* Extracts window predicate information from VEQL and gives configuration values to the window assigner to create windows for different queries. It also creates separate instances of state queues which buffers incoming states.

## VI. EVENT MATCHING

The matcher converts the query predicates to object nodes following the VEKG schema and performs event matching. As shown in Fig. 8, the matcher performs video event matching in 3 steps: 1) Object Event Matching, 2) Spatial Event Matching, and 3) Temporal Event Matching. The object and spatial matching occur within a frame (intraframe), while temporal matching can happen both within and across frames (interframe). The matching components functionality is explained below:

*Object Event Matcher:* Algorithm 1 performs object event matching and matches queries related to objects and object attributes (Q1, Q2). The object event matcher traverses object nodes from VEKG graphs and performs query matching over it. If a query object node matches with the VEKG object node, it returns it as an object event ($E_i$). For example, Fig. 8 shows the window state of 5 VEKG graphs with different object nodes. Each VEKG graph is passed to the object event matcher where

| Algorithm 1: Object Event Matching | Algorithm 2: Spatial Event Matching | Algorithm 3: Temporal Event Matching |
|---|---|---|
| begin<br>  while *State is not NULL* do<br>    $Q \leftarrow$ getQueryPredicates()<br>    for *each $VE_{KG}$ from State* do<br>      $t_g \leftarrow VE_{KG}$.getNode()<br>      $t_q \leftarrow Q$.getNode()<br>      $r \leftarrow$ Match($t_g, t_q$)<br>      if *r is TRUE*<br>        return $t_g$ as *object event ($E_i$)*<br>    end<br>  end<br>end | begin<br>  list $\leftarrow$ getObjectEventMatch()<br>  refobj $\leftarrow$ getReferenceObject(list,Q)<br>  for *each Object Event (E) from list* do<br>    $s \leftarrow$ callspatialfunction( *spatial rel, refobj, E*)<br>    if *s is TRUE*<br>      return $s$ as *spatial event ($E_i$)*<br>  end<br>end | begin<br>  $Q \leftarrow$ getQueryPredicates()<br>  $t_q \leftarrow Q$.getNode()<br>  $M \leftarrow$ createMap($t_q$, E)<br>  for *each $VE_{KG}$ from State* do<br>    $E_{tq} \leftarrow$ getObjectEventMatch()<br>    add $E_{tq}$ to map M at its $t_q$ key<br>    if $M.size() > 1$<br>      $t \leftarrow$ calltemporalfunction( $M, t_q$ )<br>      if *t is TRUE*<br>        return *t* as *temporal event ($E_i$)*<br>    end<br>  end<br>end |

it finds 'Car' query object node (Q1) in $VEKG_1$ and $VEKG_3$.

*Spatial Event Matcher:* The spatial matcher extracts the list of object nodes from a VEKG graph using the object event matcher. It extracts the reference object from the extracted list using query predicates and applies the spatial function to establish a spatial relationship between object nodes in the list (Algorithm 2). For example, in Q3, the 'Car' and 'Person' object nodes are extracted from $VEKG_3$ using the object event matcher. The spatial event matcher then sets 'Person' as a reference object, applies 'Left' boolean spatial function and updates the spatial edge weight. If the 'Left' condition satisfies then the matcher notifies it as a spatial event '$E_i$'.

*Temporal Event Matcher:* The temporal pattern matcher creates a map<key, value> where 'key' is unique and equivalent to the query object node and the value is an event object ($E_i$). For example, in Fig. 8, Q4 has two query object nodes, i.e. 'Car' and 'Person' which will be treated as keys for the map. The temporal matcher receives each object ($E_i$) from the object event matcher and puts it into the map to the corresponding key. Fig. 8 shows a map where 'Car' and 'Person' keys have value events $[E_1.t_1, E_3.t_3]$ and $[E_3.t_3, E_4.t_4]$. Here $\{t_1, t_2, t_3, t_4\}$ is the time when the event has originally occurred in the video frame and added to the VEKG graph during its construction. The temporal matcher continuously checks the size of the map and starts temporal reasoning if its size becomes greater than 1, which means two different types of object events are present. For the SEQ operation, the map keys are sorted as per their order of occurrence. We follow *skip-till-any* SEQ operation where each combination of the sequence is matched and has an exponential runtime complexity [2]. For example, in Fig. 8, $E_1.t_1 < E_3.t_3$ and $E_1.t_1 < E_4.t_4$ are two sequence patterns for Q4. For the CONJ temporal pattern, if all the map keys have an event value, then this suggests that both the required events are present irrespective of their timing order. Similarly, $E_3.t_3$ is a concurrent (EQ) temporal event pattern as both event 'Car' and 'Person' occur together at time t3.

The matcher gives an overall score to the pattern by calculating the weighted mean of all the events it has detected. In eq. 2, $P(E_i)$ is the detection probability of each object event. This is the confidence probability given by the classifier for each detected object from the image. $-log_2(P(E_i))$ is the information content or weight of each detected event. In information theory, $\sum P(E_i) * (-log_2(P(E_i)))$ is considered as the entropy, which reflects the average rate at which information is produced. For example, if P(Car) = 0.6 and P(Person) = 0.7, then M = 0.641 which is the weighted mean of both probabilities. The matcher score ($M$) is then compared with the required input confidence given by the user (Q3-confidence > 0.5). If the score satisfies the given confidence request, then it notifies that the pattern is detected.

$$Matcher_{Confscore}(M) = \frac{\sum_{i=1}^{N} P(E_i) * (-log_2(P(E_i)))}{\sum_{i=1}^{N} (-log_2(P(E_i)))} \quad (2)$$

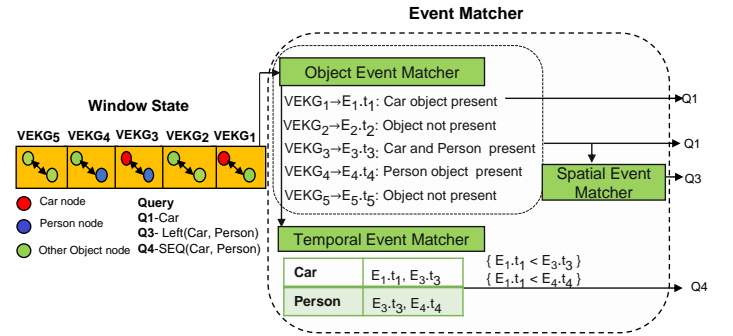

Fig. 8. Event Matching for Video Streams

VII. EXPERIMENTAL SETUP AND RESULTS

*A. Implementation and Datasets*

The VidCEP prototype is implemented in Java[1] and can handle multiple video streams and queries in parallel. All the experiments were performed on a 16-core Linux machine running on 3.1 GHz processor, Nvidia Titan Xp GPU with 12 GB of RAM. Java OpenCV library[2] was used for initial video frames decoding. For video content retrieval Deeplearning4j[3], a Java-based deep learning library was used. For object detection, we have used the DNN based Tiny YOLO [15] model. For attribute extraction, the features based on bounding box coordinates were fetched from the Tiny YOLO model layer and passed to the attribute classifier, which is a simple color filter. JGraphtT[4], a Java library for graphs was used for VEKG graph construction. We performed experiments on queries using different producers.

Table 3 shows the list of six videos collected from different datasets and websites (Pexels, YouTube), which act as producers in our system. Most of the videos are streamed at an average rate of 30 frames per second. The ground truth data for events was created manually so that it can act as a baseline for comparison. The videos were selected by visually analysing that a given pattern (like high traffic flow) is present or not. These videos cover different query dimensions which we have discussed in Section IV-A.

---

[1] Code Available at: https://github.com/piyushy1/VidCEP
[2] https://github.com/opencv-java
[3] https://deeplearning4j.org/
[4] https://jgrapht.org/

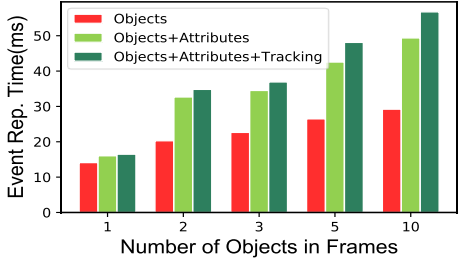
Fig. 9. Average event representation time for different video frames [38]

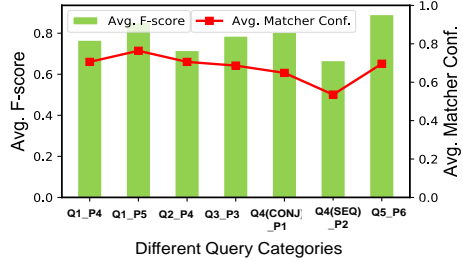
Fig. 10. Average query accuracy and matcher confidence score

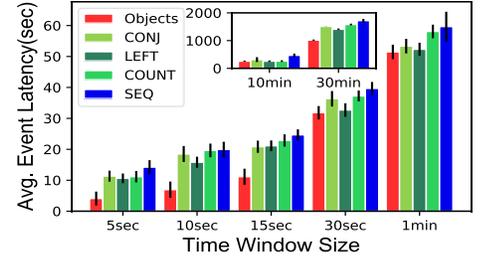
Fig. 11. Event latency (system) for different time windows with error bars

TABLE 3 DATASET SPECIFICATION

| Video | Dataset | Query | Dimensions |
|---|---|---|---|
| P1 | VidVRD [43] | Q4(CONJ) | D4 |
| P2 | Urban Tracker [44] | Q4(SEQ) | D4 |
| P3 | Pexels | Q3(Left) | D3 |
| P4 | Pexels | Q1, Q2 | D1, D2 |
| P5 | Pexels | Q1 | D1 |
| P6 | YouTube | Q5(Traffic) | D1, D4 |

### B. Evaluation Results

VidCEP is a novel video-based CEP system. To the best of our knowledge, presently there is no available baseline system for comparison.

*1) Event Representation time:* It is the time taken by the video stream processor to convert the frame to a VEKG graph [38]. Eq. 3 shows the event representation time, which includes the frame decode time to a low-level feature matrix, object and attribute classifier time and time to construct a graph from extracted labels and attributes. As the frame decoding and graph construction time was very low (0.5-1 millisecond), we have focussed only on time required by the DNN models cascade.

$$t_{event-rep} = t_{frame-decode} + t_{DNN} + t_{graph-construct} \quad (3)$$

Fig. 9 shows the average time required by a DNN model to extract relevant features for graph construction. We have focused on three key characteristics, i.e. 1) Object detection time, 2) Object and attribute detection time, and 3) Object, attribute detection and object tracking time. These three characteristics were compared with video frames having objects ranging between 1 (F1) to 10 (F2). Fig. 9 shows the average object detection time lies between 14.1 ms to 29.2 ms. The difference between object detection time is very low because of the shared computation principle over which object detectors work [15]. The object and attribute average detection time for F1 is 16.03 ms, which increases to 49.3 ms for F2. The extra overhead is because each object needs to be passed to the attribute classifier, and with an increase in object number, the attribute classification time increases. The tracking is a cheaper process, thus including tracking time, the overall detection time is 16.4 ms, and 56.7 ms for F1 and F2, respectively.

*2) Event Query Accuracy and Matcher Confidence:* Event query accuracy examines how many relevant event patterns were detected for each query as compared to the ground truth. Query accuracy is evaluated using F-score (eq. 4), which is a harmonic mean of precision and recall. Fig. 10 shows the mean F-score for different queries, which is averaged across a time window of 10 seconds.

$$F-score = \frac{2 * Precison * Recall}{Precision + Recall} \quad (4)$$

The F-score for Q1 on video producer P4(Q1_P4) is low (0.76) as compared to P5(Q1_P5) which is 0.85. The F-Score of Q1_P4 is low because there are multiple objects ('Car') in P4, which causes occlusion leading to more false positives, reducing the overall score. The F-score for Q2_P4 is 0.71, which is less as compared to Q1_P4 because of misclassifications in the attribute classifier. The F-score of Q3_P3 is 0.78, which is a spatial event pattern and is shown visually in Fig. 15. The green and red dots are the tracking points of the objects at different time. The green dots (Black Car) are on the left side of the red dots, i.e. 'Not Black Car'. Here 'Not Black Car' is treated as a reference object from which spatial direction 'Left' is calculated. We have shown two temporal relation SEQ(Q4_P2) and CONJ(Q4_P1) whose F-score are 0.66 and 0.80, respectively. Fig. 16 shows the SEQ(Q4_P2) pattern where the 'Car' event happens before the 'Person' event. In Fig. 17, a false positive instance of Q4_P2 is shown where 'Person' appears before 'Car', but the system detected it as a sequence of 'Car' and Person'. This is because in one of the frames the object detector was unable to detect 'Person' but was able to detect 'Car' and in the later frame, the system detected both 'Person' and 'Car' making it as a sequence. There can be many objects missed due to object detector limitations. Therefore SEQ has low F-score as compared to other queries. Fig. 19 shows a CONJ relation of a 'Car' and 'Person' events. Query 5 on producer 6 (Q5_P6) detects high traffic flow (Fig. 18) with F-score of 0.89.

The matcher confidence for different queries is shown as a red line in Fig. 10. The matcher confidence is associated with DNN model detection confidence [0-1], but there is a key difference. The matcher detects the overall confidence of an event pattern which may involve multiple objects, while an object detector gives a confidence score for each object. The average matcher confidence for all queries lies between 0.53 to 0.76. Thus, empirically 0.5 query confidence is a good initial cutoff for pattern detection.

*3) Matching Latency of Query Pattern:* Fig. 11 and Fig.12 show the system and matcher latency, respectively, for different queries for different window sizes. The matcher latency is the time difference between when the window state is sent to the matcher and when it notifies the pattern across that state (eq. 5). System latency is the summation of the average event representation time of window size (w) and the matcher latency (eq. 6).

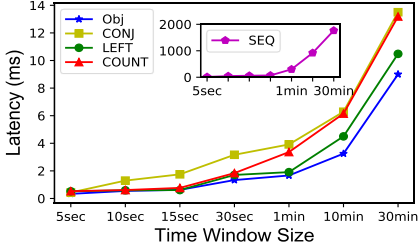
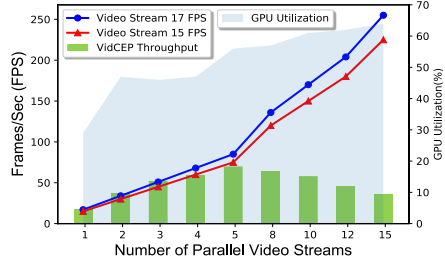
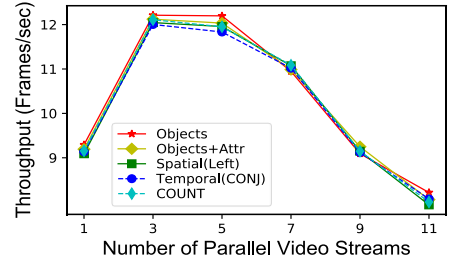

Fig. 12. Event latency (system) for different time windows with error bars

Fig. 13. VidCEP throughput (GPU)

Fig. 14. VidCEP throughput (CPU)

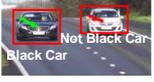
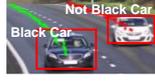
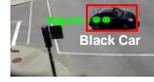
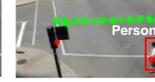
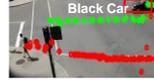

Fig. 15. Spatial relation: 'black car' left of 'not black car' (P3)

Fig. 16. Temporal relation- seq of car and person (P2)

Fig. 17. SEQ of car and person (false positive) (P2)

Fig. 18. (P6) Traffic flow

Fig. 19. CONJ of car & person(P1)

$$t_{matcher-latency} = t_{notify} - t_{window\ send\ to\ matcher} \quad (5)$$

$$t_{system-latency} = (\sum_{i=1}^{w}(t_{event-rep\ for\ window\ size(w)})/w) + t_{matcher-latency} \quad (6)$$

We ran 1-hour video for different window sizes ranging from 5 seconds to 30 mins. With an increase in the window sizes, the system latency (objects) increases (Fig. 11) ranging from 4.09 sec for the window of 5 sec to 1016.8 sec for the window of 30 mins. This happens because, with the increase in window time, it consumes more frames which require more computation for their representation and matching. Fig. 12 shows that SEQ operation has the highest matching latency (12.5 -1764.5 ms) as compared to other operators as it uses *Skip-Till-Any* method, which searches all patterns and has exponential runtime complexity. The average matcher latency for other query patterns ranges between 0.33 -13.9 ms for a window size of 5 sec and 30 mins, respectively.

*4) System Throughput:* Throughput means the number of frames the system can process per second. Fig. 13 shows that VidCEP achieves a throughput of approximately 70 frames per second (fps) (green bars) for five parallel video streams. It achieves near-real-time performance for five parallel video streams if streamed at the rate of 17 fps (blue line). Initially, the throughput of the system increases with the increase in the number of video producers. This is due to the memory availability which the system can consume. After this, the throughput starts decreasing due to memory overhead as several producers start loading the computationally intensive DNN models and videos frames in the memory. When 15 videos are streamed in parallel, VidCEP throughput decreases to 36.3 fps with 65% of GPU utilization time. Fig. 14 shows VidCEP CPU throughput performance for different operators. The system has a very low throughput of 12 fps for each query for five parallel videos streams.

## VIII. RELATED WORK

### A. Video Event Representation

Different representation techniques like ontologies [45]–[49], graphs [50] and relational tuples [30] have been proposed for multimedia events. Westermann et al. [45] proposed an 'E' event model which discussed different modelling aspects for multimedia applications. In IMGpedia [46], the authors added low-level features of the image to create a linked dataset of images, but they did not capture semantic relationships among them as done in VidCEP. In OVIS [49], the authors have developed video surveillance ontology for large volumes of videos in databases with no support for streaming. Xu et al. [47] presented a Video Structural Description (VSD) for discovering semantic concepts in the video with no CEP focus. MSSN-Onto [48] focused on event schema for multimedia sensor networks. Lee et al. [50] proposed a graphical model where nodes are represented as segmented regions (low-level visual descriptors) in an image frame and focussed on indexing while VidCEP focus is on pattern matching.

### B. Event Matching in Video Streams

Initial work by Medioni et al. [12] was focused on detecting and tracking of moving objects using low-level image features. In 'REMIND' [51], Dubba et al. used Inductive Logic Programming to match event patterns for video. Guangnan et al. proposed EventNet [52], a video event ontology with large-scale concept library based on WikiHow articles to match queries with the semantically relevant concepts. Yadav et al. [53] focused on pattern detection like 'wildfire' from the images using crowd knowledge instead of automated pattern detection. TRECVID [54] is a benchmark for evaluating content-based video event retrieval and measure their performance. Activity recognition [55], is another domain which involves detection of predefined human actions like walking, jumping, and cooking. Video analytics frameworks like NoScope [56] and FOCUS [57] provide low cost and low latency video event detection on indexed video dataset. The systems mentioned above do not focus on expressive user queries and spatiotemporal patterns. VideoStorm [9] analyzes video streams, but it does not perform pattern matching and is not publicly available.

## IX. CONCLUSION AND FUTURE WORK

In this work, we presented VidCEP, a CEP framework to detect spatiotemporal event patterns in video streams using a graph-based video event representation model Video Event Knowledge Graph (VEKG). We proposed a declarative event language (VEQL) to query video event patterns in CEP systems. The VEQL provides different spatial and temporal constructs to detect events from the video streams. VidCEP architecture was presented which details how to process video streams and perform event pattern matching using VEQL and

VEKG. The results of this study show that VidCEP can achieve near real-time performance with 70 frames per second throughput. The system achieves subsecond matching latency even for longer windows with good F-score ranging from 0.66 to 0.89 . A limitation of VidCEP is it depends on DNN models, and any prediction failure in them will decrease its performance. Future work will focus on optimizing VidCEP video processing capability and devising more expressive spatiotemporal operators.